# From A-to-Z Review of Clustering Validation Indices

Bryar A. Hassan[1,2], Noor Bahjat Tayfor[3], Alla A. Hassan[4], Aram M. Ahmed[1], Tarik A. Rashid[1], Naz N. Abdalla[5]

[1]Computer Science and Engineering Department, University of Kurdistan Hewler, Erbil, Iraq

[2]Department of Computer Science, College of Science, Charmo University, 46023 Chamchamal, Sulaimani, Iraq

[3]Department of Information Technology, Kurdistan Technical Institute, Sulaimani, Iraq

[4]Department of Database Technology, Computer Science Institute, Sulaimani Polytechnic University, Sulaimani, KRI, Iraq

[5]Information Technology Department, Faculty of Mathematics, Statistics and Computer Science, University of Tehran, Tehran, Iran

Email (Corresponding): bryar.ahmad@ukh.edu.krd

**Abstract**

Data clustering involves identifying latent similarities within a dataset and organizing them into clusters or groups. The outcomes of various clustering algorithms differ as they are susceptible to the intrinsic characteristics of the original dataset, including noise and dimensionality. The effectiveness of such clustering procedures directly impacts the homogeneity of clusters, underscoring the significance of evaluating algorithmic outcomes. Consequently, the assessment of clustering quality presents a significant and complex endeavor. A pivotal aspect affecting clustering validation is the cluster validity metric, which aids in determining the optimal number of clusters. The main goal of this study is to comprehensively review and explain the mathematical operation of internal and external cluster validity indices, but not all, to categorize these indices and to brainstorm suggestions for future advancement of clustering validation research. In addition, we review and evaluate the performance of internal and external clustering validation indices on the most common clustering algorithms, such as the evolutionary clustering algorithm star (ECA*). Finally, we suggest a classification framework for examining the functionality of both internal and external clustering validation measures regarding their ideal values, user-friendliness, responsiveness to input data, and appropriateness across various fields. This classification aids researchers in selecting the appropriate clustering validation measure to suit their specific requirements.

**Keywords**

Data Clustering Metrics, External Clustering Validation, Internal Clustering Validation, Cluster Validity Indices.

## 1. Introduction

The primary objective of clustering is to identify the inherent patterns within a collection of unlabeled data, where the items within each cluster are indistinguishable based on a specific similarity measure. Clustering is a crucial unsupervised categorization procedure that serves as a foundational component of data mining, which is widely recognized as one of the primary tasks in data analysis. Clustering is utilized in various domains, including text mining, bioinformatics, web data analysis, and data exploration. Data clustering is a sort of unsupervised learning that does not rely on any pre-existing dataset information. However, achieving favorable outcomes in clustering algorithms is contingent upon the input parameters. For example, the k-means algorithm necessitates the creation of a certain number of clusters (k) [1,2]. In this context, the inquiry is to determine the ideal quantity of clusters. Currently, there has been a growing interest in the field of cluster validity index research as a potential solution. Several cluster validities approaches have been developed in the literature that do not rely on class information [3].

Clustering validation is a method employed to identify an optimal collection of clusters that aligns most well with inherent divisions, or natural partitions, in a dataset. This methodology does not rely on any class information throughout the clustering process.

In a broad sense, clustering approaches can be categorized into three types, namely external clustering validation, internal clustering validation, and relative clustering validation.

- External clustering validation: It determines comparing the result of a cluster in terms of an externally provided label. It evaluates the purity of the cluster depending on the predefined cluster label. It could be used to find the best clustering algorithm for a particular dataset.
- Traditional internal clustering validation: It assesses the goodness of the cluster based on the underlying clustering architecture (i.e., where prior knowledge of dataset information is not required), such as the number of clusters and the clustering algorithm that has been utilized. Fundamentally, it could be applied to reveal the best clustering algorithm and the optimal number of clusters for a specific dataset.
- Newly proposed clustering validation: The recently introduced cluster validation indices refer to internal methods for validating clustering, which have been proposed in recent times.

The number of these indices differs from one source to another. According to our research from different databases (IEEE Xplore, ScienceDirect, Google Scholar, ACM Digital Library, SpringerLink, and Scopus), there are approximately forty popular validation cluster indices. Figure 1 depicts the different numbers of external and internal clustering validation indices.

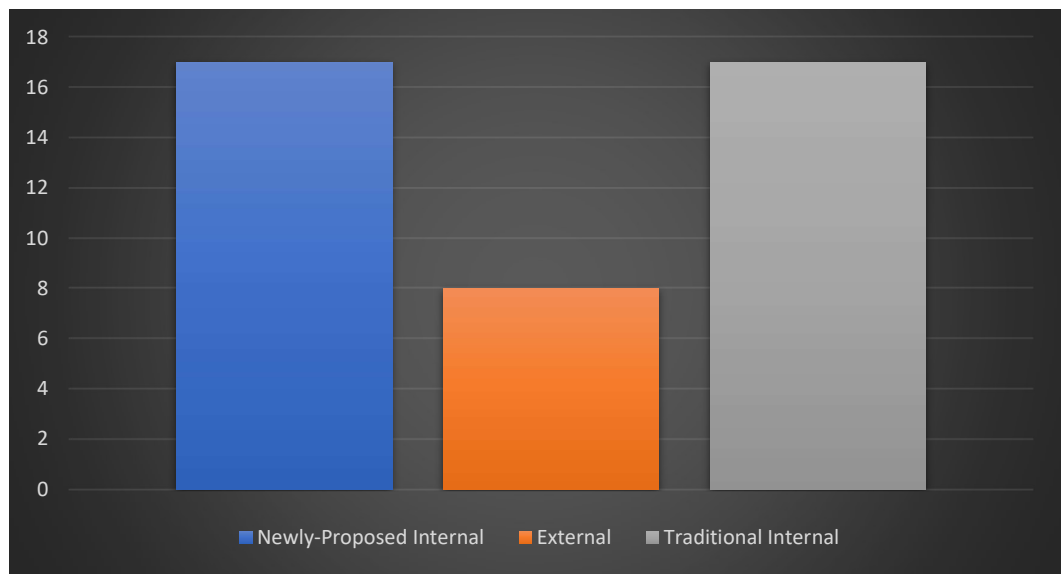

Figure 1: The growth of clustering validation indices

This review work aims to comprehensively analyze and evaluate a wide range of clustering validation metrics from a to z. This review aims to provide a detailed overview of each metric's mathematical foundation, strengths, limitations, and practical applications. By exploring the entire spectrum of available clustering validation metrics, this objective seeks to enable a deeper understanding of the nuances involved in assessing the quality and effectiveness of clustering algorithms, thereby assisting researchers, practitioners, and data scientists in making informed decisions when selecting appropriate validation metrics for their clustering tasks.

The rest of the paper is organized as follows: Section 2 presents the clustering algorithms and their validation parameters. On the other hand, Section 3 includes the most popular internal and external clustering validity indices. Furthermore, Section 4 encompasses the performance evaluation of internal and external clustering validation indices compared to the performance of the most common clustering algorithms. In addition, Section 5 proposes an operational framework to discuss the factors that significantly impact selecting the best internal and external metric. Finally, Section 6 concludes with a summary and outlines potential future work.

**2. Clustering Algorithms and Validation**

Data clustering is a tool that assists in data exploration. Some data sets have inherent categories, while others must cross the clustering process to identify specific groups. Data clustering is an unsupervised learning method that partitions a data set without prior knowledge [4]. It is widely used in various fields, such as psychology [5], biology [6], pattern recognition [7], game design [8], image processing [9], and computer security [10]. Several clustering algorithms are available for grouping datasets with various dimensions. Clustering algorithms are categorized into Partitional Clustering [11], Hierarchical Clustering [12], Density-based Clustering [13], Grid-based Clustering [14] and Model-based Clustering [15], as mentioned in Figure 2. However, the outcomes of these algorithms rely heavily on the size of the datasets, the total number of clusters in the dataset, initial conditions, and the goodness of the clustering algorithm [16,17].

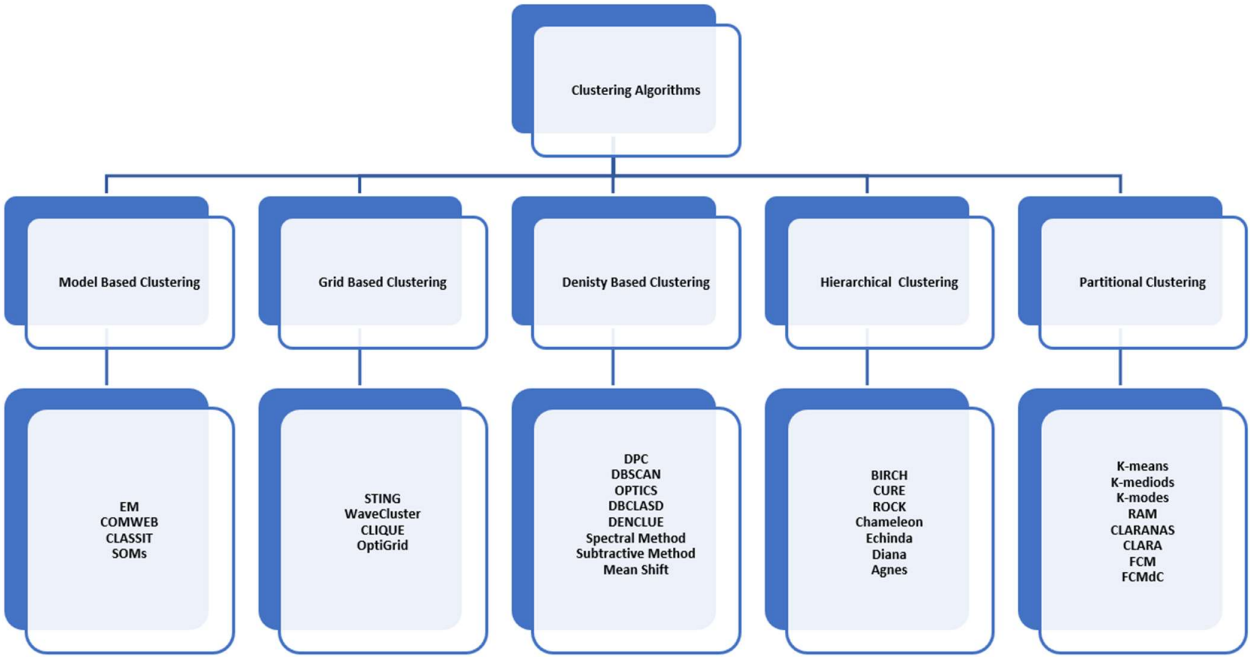

**Figure 2: Classification of clustering algorithms** [18]

One of the most popular clustering algorithms is K-means. The k-means algorithm is considered the most effective algorithm for some reasons: its implementation is very simple [19,20], its limitation is obvious compared to other algorithms, and its capacity for local fine-tuning is highly efficient [21]. As a result, it is also included in more advanced algorithms, such as the genetic algorithm [22,23], random swap [24,25], particle swarm optimization [26], density clustering [27], and spectral clustering [28]. K-means begins by selecting k points at random as initial centroids. The following two steps, assignment and update, improve this original solution. Each point is assigned to the cluster with the closest centroid in the assignment step. While in the update stage, the centroids are recomputed by calculating the meaning of each cluster's data points. These processes are repeated a predetermined number of times (called iterations) or until there is no more improvement (convergence) is met [29]. Several factors play a significant role in the performance of the k-means algorithm, such as clusters overlap, and the number of clusters; increasing dimensionality is ineffective when clusters are well separated, and strong unbalance in the clusters' size negatively affects the results [30]. However, the final number of corrected clusters is usually unknown. Creating a strategy for validating the quality of partitions after clustering is necessary as an unsupervised learning problem. It would be hard to apply various clustering findings otherwise.

The rest of this section includes the parameters of clustering validation and the objective function of clustering validation metrics.

**2.1 Parameters of Clustering Validation**

The previous clustering validity sorts rely firmly on the following four criteria:

**A. Compactness:** Internal cohesiveness is measured using the squared error function, the sum of squares within (SSW), or the sum of squares error (SSE). Thus, it determines the degree of closeness between the data points. It simply refers to the total of the squared distances between each point in the cluster and its centroid. The partition is indicated as good if the value of its variance is low. Therefore, the value of the compactness measure is minimized [31].

**B. Separation:** The external separation is quantified using the sum of squares between (SSB). Where it assumes the distances between separate clusters, typically, it is measured as the sum of the squared distance between the global average point and its centroid. A high value determines separation with high quality. Therefore, the value of the separation measure is maximized [32]. Although, the K-means clustering algorithm does not perform perfectly when the clusters are well-separated [30].

**C. Exclusiveness:** It refers to the tendency of the objects to be clustered based on the mean value [33].

**D. Incorrectness:** It indicates the percentage of risk factors [33].

As mentioned in Table 1, a good partition will have short Intra-Cluster distances and big Inter-Cluster distances as a rule [34]. However, those indices that depend on geometric-based theory do not perform well on the noisy dataset. Therefore, the two new criteria, exclusiveness and incorrectness, rely fundamentally on the statistical-based theory, which has been proposed. Where Exclusiveness aims to maximize the exclusion of outliers because their presence gives us invalid outcomes, and Incorrectness seeks to minimize the percentage of the loss function that results in improperly projecting data to the cluster [33].

Table 1: The Properties of Cluster Validity Indices (CVIs) [33]

| No. | Criteria | Measuring Principle | Definition | Dependability |
|---|---|---|---|---|
| 1 | Compactness | Intra-Cluster Distance | It minimizes the sum of distances of all the points and the centroids within the same cluster. | Geometrical-Based Theory |
| 2 | Separability | Inter-Cluster Distance | It maximizes the sum of the squared distance between all clusters. | Geometrical-Based Theory |
| 3 | Exclusiveness | Probability Density Function | It applies the Gaussian Normal Distribution function to maximize the outliers, avoiding the problem of clustering points depending on the mean. | Statistical-Based Theory |
| 4 | Incorrectness | Loss Function | It measures the median of each cluster, and it applies the Loss Function to minimize the risk proportion. | Statistical-Based Theory |

**2.2 Objective Function**

It is also called sum-of-squared errors (SSE) [30]Whether the factor combination of CVIs is minimized or maximized, the most appropriate number of clusters could be obtained. The best CVI will be defined by minimizing the objective function.

Objective Function (OBF) = Min (Compactness) + Max (Separability) + Max (Exclusiveness) + Min (Incorrectness) [33]

OBF defines mathematically every single cluster in the n-dimensional dataset:

$$\text{OBF} = \left\{ Min \left[ C = \sum_{i=1}^{n} \sum_{j=1, j \neq i}^{n} \|x_{ij} - y_{ij}\|^2 \right] + Max \left[ S = \sum_{i=1}^{n} \sum_{j=1, j \neq i}^{n} \|c_i - c_j\|^2 \right] + Max \left[ S = \sum_{i=1}^{n} \sum_{j=1, j \neq i}^{n} \|c_i - c_j\|^2 \right] + Max \left[ \frac{1}{(2\pi)^{\frac{k}{2}} |\Sigma|^{\frac{1}{2}}} e^{\frac{1}{2}(x - \mu)' \Sigma^{-1} (x - \mu)} \right] + Min[I = E(L(x, \mu))] \right\} \quad (1)$$

**3. Types of Clustering Validation Indices**

Clustering validation types encompass a range of techniques to assess the quality and effectiveness of clustering algorithms. These methods play a crucial role in evaluating the resulting clusters, determining the optimal number of clusters, and providing insights into the coherence and separation of data points within clusters. These validation types can be broadly categorized into internal, external, and relative indices. Internal indices focus on the inherent structure of the data and the clustering results in themselves, evaluating factors like intra-cluster cohesion and inter-cluster separation. External indices, on the other hand, leverage known ground truth labels to measure the agreement between the clustering and the true classes. Newly proposed clustering validation refers to internal methods for validating clustering, which have been proposed in recent times. Figure 3 summarizes the types of clustering validation indices.

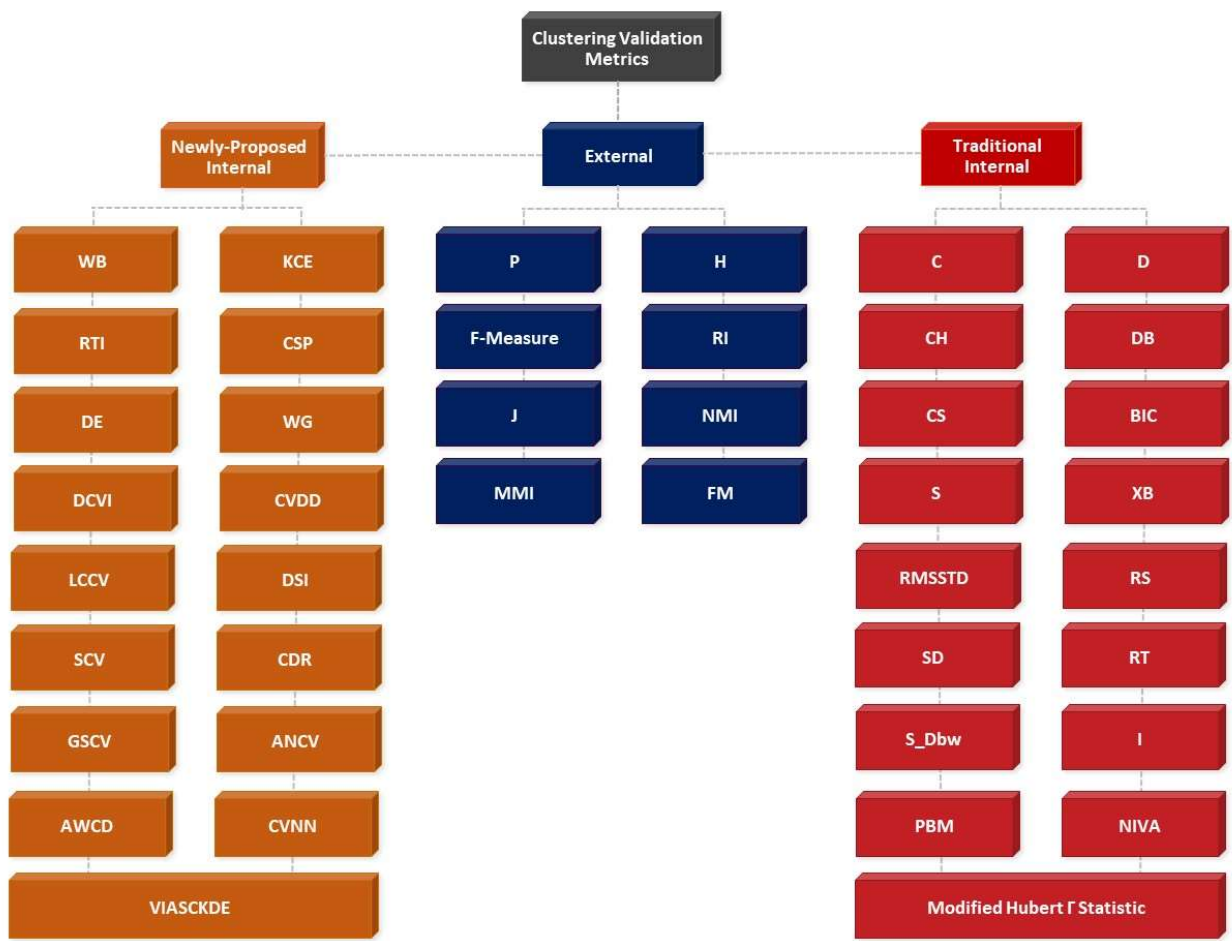

Figure 3: types of clustering validation indices

**3.1 Internal Clustering Validation Indices**

This section describes seventeen most popular traditional internal clustering validation indices, which are: C Index (C), Dunn Index (D), Calinski-Harabasz Index (CH), Davies-Bouldin Index (DB), Modified Hubert Statistics Index (Γ), Bayesian Information Criterion Index (BIC), Silhouette Index (S), Xie-Beni Index (XB), Root-mean squared Standard Deviation (RMSSTD), R-Squared (RS), Standard Deviation Index (SD), Ray-Turi Index (RT), S_Dbw Index, I Index (I), CS Index, PBM Index and NIVA Index. Table A.1 depicts each internal clustering validity index and compares them in terms of optimal value, runtime complexity, pros and cons, and the applications that use those indices in their fields.

**3.2 Newly Proposed Clustering Validation Indices**

This section describes the seventeen most popular internal clustering validation indices, which are: Clustering Validation based on Nearest Neighbor (CVNN), WB Index, KCE, Representative Tree Index (RTI), Compact-Separate Proportion Index (CSP), Density Estimation Index (DE), Wemmert Gancarski Index (WG), Density-Core-based Clustering Validation Index (DCVI), Clustering Validity Index based on Density-Involved Distance (CVDD), Local Cores-based Cluster Validity Index (LCCV), Distance-based Separability Measure (DSI), Synthetical Clustering Validity (SCV), Generalized Synthetical Clustering Validity (GSCV), Adjustment of within-cluster Distance (AWCD), Contiguous Density Region (CDR) Index, Validity Index for Arbitrary-Shaped Clusters based on the Kernel Density Estimation (VIASCKDE) and Augmented Non-shared Nearest Neighbors (ANCV). Table B.1 depicts each internal clustering validity index that was proposed after 2014 and compares them in terms of optimal value, runtime complexity, pros and cons, and the applications that use those indices in their fields.

### 3.3. External Clustering Validation Indices

This section describes the eight most popular external clustering validation indices, which are: Purity (P), Entropy (H), F-Measure (F), Rand Index (R), Jaccard Index (J), Normalized Mutual Information (NMI), Min-Max Index (MMI), Fowlkes-Mallows Index (FM). Table C.1 represents each external clustering validity index and compares them in terms of optimal value, runtime complexity, pros, and cons, and the applications that use those indices in their fields.

## 4. Comparative evaluation

This section includes a performance evaluation of internal and external clustering validation indices versus the performance of the most common clustering algorithms.

### 4.1 Performance Evaluation of Clustering Validation Indices

This section experiments with the performance evaluation of clustering validation metrics using ECA* as one of the recent, well-known, and well-performed clustering algorithms [35]. For this purpose, we utilize 12 datasets that previously have been used by Chien-Hsing Chou [36][37], and Maria Halkidi [38].

Table 2 represents the total number of correctly identified clusters using five internal validity indices (C, D, CH, DB, Γ, BIC, S, XB, RMSSTD, RS, SD, RT, S_Dbw Index, I, CS, PBM, NIVA, and PDBI) after employing the ECA*. C, D, Γ, S, XB, and PBM, RMSSTD, RS, SD, RT, CS, PBM, C, and PBM Indices obtained the best results after employing ECA*. On the one hand, CS gets the highest correct clusters after utilizing the ECA*, followed by C and PBM. On the other hand, I, S_Dbw Index, NIVA, and PDBI identify the least correct number of clusters for most of the datasets.

Table 2: A summary of the outcomes produced by applying internal validity indices to the ECA*

| Internal Validity Indices/Datasets | 1 | 2 | 3 | 4 | 4 | 5 | 6 | 7 | 8 | 9 | 10 | 11 | 12 | CMC | Total (-) | Total (+) |
|---|---|---|---|---|---|---|---|---|---|---|---|---|---|---|---|---|
| C | - | + | + | + | + | + | + | - | - | - | + | + | + | + | 4 | 10 |
| D | - | + | - | - | + | - | - | + | - | + | + | + | - | + | 7 | 7 |
| CH | - | - | + | + | + | - | + | + | - | - | - | + | - | - | 8 | 6 |
| DB | - | + | - | - | - | + | + | - | - | - | - | + | + | - | 9 | 5 |
| Γ | + | - | - | + | + | - | - | - | + | - | + | - | + | + | 7 | 7 |
| BIC | + | - | - | + | + | - | - | - | + | - | - | + | + | - | 8 | 6 |
| S | + | + | + | + | + | - | + | - | + | - | + | - | - | - | 6 | 8 |
| XB | - | - | + | + | - | + | - | - | + | + | + | - | - | + | 7 | 7 |
| RMSSTD | + | + | + | - | + | + | - | - | - | + | + | + | - | - | 6 | 8 |
| RS | + | - | + | - | - | + | - | + | + | - | + | - | + | - | 7 | 7 |
| SD | + | - | + | + | + | + | + | + | - | - | + | - | - | - | 6 | 8 |
| RT | + | - | + | + | - | - | - | + | + | + | + | - | - | - | 7 | 7 |
| S_Dbw Index | - | + | - | - | - | - | - | - | + | + | + | - | + | - | 9 | 5 |
| I | - | - | - | - | - | + | - | - | - | + | + | - | + | + | 9 | 5 |
| CS | - | + | + | + | + | + | + | + | + | + | + | + | + | + | 1 | 13 |
| PBM | + | - | - | + | - | + | + | + | + | - | + | + | + | + | 4 | 10 |
| NIVA | - | - | - | + | - | - | + | - | - | + | + | + | + | - | 8 | 6 |
| PDBI | + | - | - | - | - | + | - | - | - | - | + | + | + | + | 8 | 6 |
| Average | | | | | | | | | | | | | | | 6.722 | 7.277 |

Furthermore, Table 3 shows a summary of the outcomes produced by applying newly proposed validity indices to the ECA*. From where we can see that the WB index correctly identifies the number of clusters in all trials, and ANCV finds the correct number of groups in 11 datasets, while CSP only gives correct results in Dataset 8. Meanwhile, WG and GSCV find the correct cluster in half of the datasets.

Table 3: A summary of the outcomes produced by applying newly proposed validity indices to the ECA*

| Newly Proposed Internal Validity Indices/ Datasets | 1 | 2 | 3 | 4 | 4 | 5 | 6 | 7 | 8 | 9 | 10 | 11 | 12 | CMC | Total (-) | Total (+) |
|---|---|---|---|---|---|---|---|---|---|---|---|---|---|---|---|---|
| CVNN | - | - | + | + | - | + | - | + | + | - | - | - | - | + | 8 | 6 |
| WB Index | + | + | + | + | + | + | + | + | + | + | + | + | + | + | 0 | 14 |
| KCE | - | - | - | - | - | - | - | + | + | + | - | - | + | + | 9 | 5 |
| RTI | + | - | - | - | + | - | - | + | - | - | - | - | + | - | 10 | 4 |
| CSP | - | - | - | - | - | - | - | - | + | - | - | - | - | - | 13 | 1 |
| Density Estimation Index | + | + | - | + | + | - | - | - | - | - | - | - | + | - | 9 | 5 |
| DE | - | + | - | + | - | - | + | - | - | + | - | - | - | - | 10 | 4 |
| WG | + | - | - | + | - | + | + | - | - | + | + | + | - | - | 7 | 7 |
| DCVI | - | - | + | - | - | - | - | + | - | + | + | - | + | + | 8 | 6 |
| CVDD | + | - | - | + | + | - | - | + | + | + | - | + | - | + | 6 | 8 |
| LCCV | + | + | - | - | - | + | + | - | + | + | + | + | - | - | 6 | 8 |
| DSI | + | + | - | - | - | + | + | + | + | + | + | - | - | - | 6 | 8 |
| SCV | - | + | - | - | - | + | + | + | + | - | + | + | + | - | 6 | 8 |
| GSCV | - | - | + | - | - | + | - | + | - | + | - | + | + | + | 7 | 7 |
| AWCD | - | + | + | - | + | + | - | + | + | - | - | + | + | - | 6 | 8 |
| CDR | + | - | - | - | + | + | + | + | - | + | + | - | - | + | 6 | 8 |
| VIASCKDE | + | - | - | - | - | + | - | + | - | - | - | - | - | + | 10 | 4 |
| ANCA | + | + | + | + | + | + | - | + | + | - | + | + | + | - | 3 | 11 |
| Average | | | | | | | | | | | | | | | 7.222 | 6.777 |

Lastly, Table 4 summarizes the outcomes produced by applying external validity indices to the ECA*. F, R, J, and FM give better results when ECA* is used. J obtains the most accurate results in 10 out of 14 trials, whereas MMI achieves the least accurate results in 8 out of 14 trials. Also, P, H NMI identifies only 6 correct clusters out of 14.

Table 4: A summary of the outcomes produced by applying external validity indices to the ECA*

| Internal Validity Indices/ Datasets | 1 | 2 | 3 | 4 | 4 | 5 | 6 | 7 | 8 | 9 | 10 | 11 | 12 | CMC | Total (-) | Total (+) |
|---|---|---|---|---|---|---|---|---|---|---|---|---|---|---|---|---|
| P | + | + | - | + | + | + | - | - | - | - | - | + | - | - | 8 | 6 |
| H | - | + | - | + | + | + | - | + | - | - | + | - | - | - | 8 | 6 |
| F | + | - | - | + | - | + | - | + | + | + | + | - | + | + | 5 | 9 |
| R | + | - | + | + | + | - | + | + | - | - | - | + | + | + | 5 | 9 |
| J | - | + | + | + | + | - | + | - | - | + | + | + | + | + | 4 | 10 |
| NMI | + | + | - | + | - | - | - | + | - | + | + | - | - | - | 8 | 6 |
| MMI | + | + | - | - | - | + | - | - | - | + | - | - | + | - | 9 | 5 |
| FM | + | + | + | - | + | - | - | - | + | - | - | + | + | + | 6 | 8 |
| P | + | - | + | - | + | + | - | + | + | - | + | - | - | - | 7 | 7 |
| Average | | | | | | | | | | | | | | | 6.66 | 7.33 |

In conclusion, from these tests, it can be said that external validity indexes are more accurate in the identification of correct cluster numbers in groups formed with the ECA*.

**4.2 Performance Evaluation**

This sub-section reveals the related works carried out on performance evaluation of clustering indices.

The performance of four popular clustering algorithms (K-Means, Bisecting K-Means, Fuzzy C-Means, and Genetic K-Means) has been compared on 20 real-life datasets from various domains. The evaluation is based on three internal validity indices (Silhouette, Dunn, and DB) and three external validity indices (Rand, Jaccard, and FM). The results show that Genetic K-Means performs the best overall, while Fuzzy C-Means performs better when there are overlapping clusters. It has also been suggested further improvement of the performance of Genetic K-Means and considering cluster density along with distance measure in a multi-objective Genetic K-Means framework [39].

Internal validity measures have been investigated by comparing six kinds of clustering algorithms (i.e., K-means, DBSCAN, Spectral, Ward, Meanshift, and EM). Four clustering validation indices have been utilized to compare the performance of clustering algorithms. The results showed that none of the four internal validity measures can be used to make a fair comparison between the six clustering algorithms tested. All measures exhibit some undesired properties, such as sensitivity to points identified as noise, a preference for highly imbalanced solutions, or a bias toward spherical clustering. The lack of an appropriate performance measure is the main obstacle to applying techniques for meta-learning and model selection from supervised learning to clustering. However, it has been recommended that using k-means is sufficient to achieve high ratings on the SI and CH measures. In addition, it has been found that comparing clustering algorithms using external criteria can be misleading and that highly imbalanced clustering tends to score well on internal validity measures [40].

Another study discussed using internal clustering validation indices (CVIs) to measure the quality of clustering algorithms like K-means, K-medians, and K-spatial medians. The performance of several CVIs with different distance measures and clustering algorithms have been evaluated and compared on synthetic datasets (such as A1, S-sets, Dim-sets, Birch2, Unbalance, and G2) and real datasets (for instance, Iris, Arrhythmia, Steel Plates, Ionosphere, and USPS). In terms of comparing CVIs, the WG index showed consistent behavior with real datasets and beat other indices in tests with synthetic data. Except for the Sim5 dataset, which had clusters of varying sizes and overlaps, it correctly determined the total number of clusters across all other datasets. In addition to WB KCE and CH for robust clustering and K-means, PBM was an index with good and steady performance. Only WB and KCE for K-means and PBM with robust estimates found the correct number of clusters in the Sim5 data. These results indicate that DB and RT should be avoided. Some indices (CH, DB, RT) tend to perform worse and fail more frequently as the number of clusters increases, and there appears to be a correlation between the two. Concerning comparing distance metrics (i.e., Squared Euclidean distance, Euclidean distance, and City-Block) and clustering methods, the WG index emerges as the best. Both WB and KCE do exceptionally well across the board. Some distances perform better than others when using specific indices; for instance, the squared Euclidean distance is ideal for CH, whereas city-block and Euclidean distances are optimal for PBM. Regarding the original K-means, WG, KCE, and CH are the three best indices to utilize; regarding the robust variations with K-medians and K-spatial medians, WG, PBM, and WB have the highest success rate. The outcomes show that no single CVI has a clear advantage in every context, but each is best suited to a certain kind of data. Furthermore, the characteristics of these indices with multiple datasets and different clustering algorithms were evaluated [41].

Several validity indices have been compared to determine clustering algorithms' optimal number of clusters. The K-means algorithm is used over a range of cluster numbers, and the best indices values are extracted to determine the optimal number of clusters. The effectiveness of these indices is tested on synthetic and satellite datasets. The results show that the Sum of Squares index is effective among a large set of data. Moreover, a comparative study of different validity indices for clustering algorithms has been presented using synthetic, composite, and real data. The WB index is found to be the most effective in extracting the optimal number of clusters. It also demonstrated the adaptability of the tested indices to different overlapping dimensions and complexity rates. The results suggest the potential for developing a completely automated clustering method using the best index in combination with an improved clustering algorithm [42].

Nine clustering algorithms (subspace, EM, spectral, Clara, model, k-means, hierarchical, optics, and dbscan) have been compared in R language using 400 distinct artificial datasets for education with different sizes and features. Seven clustering algorithms have been utilized to conduct the comparison study: K-mean, K-medoids, Expectation-maximization (EM), Agglomerative clustering, Spectral clustering, DBSCAN, and OPTICS. For evaluating the clustering algorithms mentioned above, internal, and external indices have been utilized. The results obtained depend mainly on the K value in the clustering algorithm. When the value of K<10, the performance is significantly low, while when the value of K>10 increases, the performance of the clustering algorithms. The results of Silhouette and Dunn indices vary when selecting different numbers of features and classes. Dunn index shows lower outcomes as it incorrectly estimates the cluster numbers for the hierarchal algorithm. Although, a higher accuracy has been obtained when the number of classes about K=10. Lastly, the outcomes indicate that spectral clustering methods perform well (compared to others) when the default configuration of clustering methods is considered and the external evaluation metrics such as Adjusted Rand Index (ARI) and Jaccard Index (JI), Fowlkes-Mallows Index (FM) and Normalized Mutual Information Index (NMI) have been utilized to assess the goodness of the partitions obtained by each algorithm [43].

An automatic comparative approach has been proposed for evaluating clustering algorithms in an educational context. The approach uses multiple performance metrics and the TOPSIS [44,45] method to rank the algorithms. Moreover, the effect of data normalization has been investigated on the performance of the algorithms. The experiments are conducted on datasets extracted from the Moodle system of the University of Tartu in Estonia. The results show that spectral clustering is preferred in datasets with a small and medium number of features, while OPTICS is the best option for datasets with many features. Normalization hurts the performance of spectral clustering but mostly improves the performance of other methods. The study evaluates the performance of different clustering algorithms in an educational context and models the evaluation task as a multiple-criteria decision-making problem. The TOPSIS method is employed to rank the results produced by multiple internal and external performance measures when applied to several clustering algorithms. The study finds that spectral clustering is more robust and outperforms other clustering methods for datasets with 15 or fewer features, while OPTICS strongly outperforms other methods for datasets with many features. The study also examines the effect of normalization on the quality of the clusters produced by different clustering methods. Finally, it has been concluded that the proposed approach can help educators and researchers select the most appropriate clustering algorithm for their specific needs [46].

UCI program LIBRAS [47] dataset has been utilized to evaluate the performance of K-mean and hierarchical algorithms via utilizing internal and external indices. More specifically, two clustering methods have been compared by calculating their separation using the Pearson relationship rather than the more conventional Euclidean and squared Euclidean correlations, which most authors use but do not provide a consistent separation capability. The final decision was that k-means is preferable to hierarchical computation when considering internal and external indices [48].

## 5. Classification of Using Clustering Metrics

In this section, we propose a classification framework to investigate the operation of internal and external clustering validation indices in terms of their optimal values, easiness, or difficulty to use, sensitivity towards input data, and their application in different domains. This helps researchers to choose a proper clustering validation index for their needs. However, all CVIs have positive and negative sides. It is still difficult to recognize the best metric that could determine the correct number of clusters because it mainly depends on the clustering algorithm, the distance metric [49], the shape of the cluster, the number of clusters, and the kind of dataset that is used. Moreover, some impacts might be a significant challenge in selecting the accurate internal validation metric: monotonicity, noise, density, the existence of sub-clusters, skewed distributions the datasets with arbitrary shapes [50], cluster overlap, dimensionality, and cluster similarity measure [51]. Additionally, shared nearest neighbors and non-shared nearest neighbors could play a significant role in choosing the optimal CVI and obtaining the correct number of clusters [52]. Furthermore, internal validation indices have varying degrees of stability concerning missing data and their performance could be improved by utilizing the modified version of internal validation indices in the presence of missing data [53]. Therefore, it is highly recommended to utilize multiple indices when doing cluster analysis [41].

Moreover, external cluster validity indices as mentioned earlier are measures that are used to evaluate the quality of candidate partitions by comparing them with a ground truth partition. Although those external metrics have several advantages and disadvantages, they are used to determine how well a clustering algorithm has performed in grouping similar data points and separating dissimilar ones. The challenges of external cluster validity indices include the presence of bias, which can affect the interpretation of clustering validation results. Several types of bias can affect external cluster validity indices, including NC bias, GT bias, GT1 bias, and GT2 bias. NC bias arises when the mathematical model of the external CVI tends to be monotonic in the number of clusters in candidate partitions. GT bias arises when the ground truth partitions alter the NC bias status of an external CVI. GT1 bias arises when the numbers of clusters in the ground truth partitions alter the NC bias status of an external CVI, and GT2 bias arises when the distribution of the ground truth partitions alters the NC bias status of an external CVI. Another challenge is the selection of an appropriate external CVI, as different CVIs may be more suitable for different types of data and clustering algorithms. Finally, the performance of external cluster validity indices may be affected by the quality of the ground truth partition, as well as the size and complexity of the data set [54].

Also, the number of clusters could be fundamentally determined by the external validity indices. Moreover, it has been proven that the stability-based approach is a method for determining the number of clusters in clustering algorithms. It involves adding randomness, cross-validation strategy and normalization, selection of the external index, and selection of the clustering algorithm. A clustering solution is defined as stable if it remains the same when applied to several datasets generated with the same process or from the same underlying model. The similarity between every two clustering solutions is measured by an external validity index. The most stable result is achieved when the correct number of clusters is applied. However, stability can also be achieved with a smaller number of clusters when the positioning of clusters is not symmetrical [55].

Lastly, the performance of external validity indices might decrease when a highly imbalanced dataset is used. To solve this issue, more systematic experiments should be conducted to investigate the sensitivity levels displayed by the indices and characterize the trade-off that may exist among the desired properties [56].

## 6. Conclusion

In this article, we comprehensively assessed and elucidated the mathematical operations behind internal and external cluster validity indices, though not all of them. The aim was to categorize these indices and generate ideas for the future advancement of research in clustering validation. Additionally, the performance of internal and external clustering validation indices on widely used clustering algorithms like ECA* was reviewed and evaluated. Finally, a practical classification method was proposed for the evaluation of the functionality of both internal and external clustering validation measures concerning their ideal values, ease of use, responsiveness to input data, and suitability across different fields. This classification assisted researchers in choosing the appropriate clustering validation measure based on their specific needs. According to the experiment carried out for the performance evaluation of clustering validation metrics using ECA* for internal, newly proposed, and external clustering validation indices, it can be said that external validity indexes are more accurate in the identification of correct cluster numbers in groups formed with the ECA*. Furthermore, the categorization system to study internal and external clustering validation indices' ideal values, usability, sensitivity to input data, and fields of use was proposed. We believe that this helps researchers choose a required clustering validation index. In the future, we aim to obtain a new dataset, try to apply all the metrics that have been reviewed so far and compare and assess their results against the research that has been published. For future reading, the authors advise the reader could optionally read the following research works [35,57–73].

**Acknowledgments:** The authors thank their institutions for providing the resources and ongoing assistance necessary to undertake this research.

**Conflict of interest:** The authors declare no conflict of interest to any party.

**Ethical Approval:** The manuscript is conducted in the ethical manner advised by the targeted journal.

**Consent to Participate:** Not applicable.

**Consent to Publish:** The research is scientifically consented to be published.

**Funding:** The research received no funds.

**Competing Interests:** The authors declare no conflict of interest.

**Availability of data and materials:** Data can be shared upon request from the corresponding author.

# Appendix A

This appendix depicts Table A.1 for each internal clustering validity index and compares them in terms of optimal value, runtime complexity, pros, and cons, and the applications that use those indices in their fields.

Table A.1: The summary of the traditional internal clustering validation methods

| No. | Index Name, Reference | Optimal Value | Runtime Complexity | Advantages | Disadvantages | Applications |
|---|---|---|---|---|---|---|
| 1 | C Index (C), [74][75]. | Max | $O(n^2)$ Where: n refers to the cluster number. | 1. It is easy to compute and interpret. 2. It is fast and efficient in terms of computational time. 3. It does not require any prior knowledge about the data or the number of clusters. 4. It works well with datasets of various shapes and sizes. 5. It can handle outliers well. | 1. It can only be used for evaluating the validity of clustering results obtained using the k-means algorithm. 2. It assumes that the clusters are spherical, equally sized, and have the same density. 3. It is sensitive to the scale of the data, so normalization may be necessary. 4. It can be affected by the presence of noise or overlapping clusters. 5 It does not consider the clustering structure of the data. | Data Mining [76,77], Image Segmentation, Bioinformatics, and Text Classification. |
| 2 | Dunn Index (D), [78][79]. | Max | $O(n^2)$ Where: n refers to the data points. | 1. It is a simple and intuitive index that is easy to understand and implement. 2. It can handle clusters of different sizes and shapes effectively. 3. It is independent of the data distribution and can be used with various distance metrics. 4. It is a relative index, which means that it compares cluster compactness and separation against each other, rather than against an external criterion. 5. It can be used to estimate the optimal number of clusters by comparing the index values across different clustering solutions. | 1. It is sensitive to the scale of the data and requires normalization or standardization of the variables. 2. It assumes that clusters are well-separated, which may not be the case in high-dimensional or noisy datasets. 3. It may not be suitable for sparse or categorical data, as it relies on meaningful distances between data points. 4. It may not be robust to outliers, as they can affect the calculation of cluster compactness and separation. 5. It does not provide any information on the underlying structure or interpretation of the clusters and should be used in conjunction with other validation methods and domain knowledge. | Data Mining, Image Segmentation, Bioinformatics, and Information Retrieval. |

| | | | | | | | |
|---|---|---|---|---|---|---|---|
| 3 | Calinskie-Harabasz Index (CH), [80]. | Max | O(n) Where: n refers to the data points. | 1. It is relatively easy to calculate and interpret. 2. It doesn't require the true labels of the data since it is based solely on the data and the clustering results. 3. It favors clusters that are dense and well-separated, which is often desirable. | 1. It does not work well for datasets with highly irregular shapes or sizes, where other clustering validation indices might be more appropriate. 2. It assumes that clusters are spherical and equally sized, which is often not the case in real-world datasets. 3. The index can be heavily influenced by outliers, which can skew the results. | Image Segmentation, Customer Segmentation in marketing and Bioinformatics, and Anomaly Detection in network traffic. |
| 4 | Davies-Bouldin Index (DB), [81][82]. | Min | O(n) Where: n refers to the data points. | 1. Intuitive Interpretation: The index compares the compactness and separation properties of a clustering solution, which are both important aspects in evaluating the quality of a clustering algorithm. 2. No Assumption of Spherical Shapes: Unlike some other clustering validity metrics such as the Silhouette coefficient, the Davies-Bouldin index does not assume that the clusters are spherical or even have a uniform density. 3. Efficient Computation: The Davies-Bouldin index is relatively straightforward to compute, requiring only the calculation of distance measures between the cluster centroids. | 1. Sensitive to Input Parameters: The Davies-Bouldin index is sensitive to the choice of the number of clusters and may produce inconsistent results if the clustering algorithm is sensitive to the initial conditions or the choice of the distance metric. 2. Limited to Euclidean Distance: The Davies-Bouldin index is sensitive to the choice of distance metric and may not work well with non-Euclidean distance measures. 3. May Struggle with Overlapping Clusters: The Davies-Bouldin index may struggle to evaluate clustering solutions with overlapping clusters, as it assumes that the clusters are disjoint. | Image Segmentation, Document Clustering, Gene Expression Data Analysis, Market Segmentation, Feature Selection and Outlier Detection. |

| 5 | Modified Hubert Statistics Index (Γ), [83]. | Elbow | $O(n^2)$ Where: n refers to the data points | 1. It is a simple and easy-to-understand metric for measuring the quality of clustering results. 2. It is computationally efficient for small to medium-sized datasets. 3. It is sensitive to the shape of the clusters, unlike some other clustering validity indices. 4. It does not require knowledge of the true or optimal clustering solution. | 1. It is sensitive to noise and outliers, which can result in misleading results. 2. It does not consider the number of clusters in the dataset, which can make it difficult to compare results across different datasets with varying numbers of clusters. 3. It assumes that clusters are spherical and evenly sized, which may not be true in all cases. 4. It can produce inconsistent results for datasets with varying densities. | Image Segmentation, Pattern Recognition, Data Mining, Bioinformatics and Gene Expression Analyzing. |
|---|---|---|---|---|---|---|
| 6 | Bayesian Information Criterion Index (BIC), [84]. | Min | $O(n)$ Where: n refers to the data points. | 1. Simplicity: BIC is a relatively simple formula that is straightforward to calculate. 2. Model Selection: BIC can be used to select the correct number of clusters in a clustering algorithm by comparing the value of BIC across different numbers of clusters. 3. Account for model complexity: BIC accounts for model complexity in its calculation. It penalizes models that are too complex and favors models that are simpler yet still have a high fit. | 1. Assumption: BIC assumes that the data is generated from a normal distribution. It may not work well for data that does not follow this assumption. 2. Sensitivity to sample size: BIC can be sensitive to sample size. small sample sizes can lead to overfitting or underfitting of the model. 3. Uncertainty: BIC assumes that all models being compared are the true models generating the data. This may not always be the case and can lead to uncertainty in the selected model. | Model Selection, Clustering Analysis, Gene Expression Analysis, Image Segmentation and Text Clustering. |
| 7 | Silhouette Index (S), [85]. | Max | $O(n^2)$ Where: n refers to the data points. | 1. Silhouette index is simple to calculate and requires no additional parameters. 2. It can be used to evaluate the quality of a clustering solution and quickly identify problematic cases. | 1. Silhouette index may not work well with datasets that have noise or outliers. 2. It requires the use of a distance metric, which may not work well with all types of data. 3. In large datasets, the computation of pairwise distances between all data | Image Segmentation, Customer Segmentation in marketing, Facial Recognition, and Genetic Clustering. |

| | | | | 3. It does not make any assumptions about the shape or size of a cluster, making it suitable for non-spherical clusters.<br>4. It provides a score for each data point, which can be useful in identifying outlier points. | points can be time-consuming.<br>4. It may not be suitable for selecting the optimal number of clusters as it is sensitive to the number of clusters and may not give the same results for different cluster numbers. | |
|---|---|---|---|---|---|---|
| 8 | Xie-Beni Index (XB), [86]. | Min | $O(n^2)$<br>Where:<br>n refers to the data points. | 1. Xie-Beni is a fast and simple clustering validity index.<br>2. It's based on the compactness and separation of clusters, which provides a useful measure of the quality of the clustering.<br>3. It can be used to evaluate the performance of various clustering algorithms.<br>4. It doesn't require knowledge of the ground truth labels. | 1. Xie-Beni relies on the Euclidean distance metric, which may not be suitable for all types of data.<br>2. It's sensitive to the number of clusters and the initialization of the centroids.<br>3. It doesn't consider the shape of the clusters or the density of the data points.<br>4. It may not be appropriate for high-dimensional datasets as the curse of dimensionality can affect the performance of the clustering algorithms. | Data Mining, Image Segmentation, Pattern Recognition, and Anomaly Detection System. |
| 9 | Root-mean squared Standard Deviation (RMSSTD), [87]. | Elbow | $O(n)$<br>Where:<br>n refers to the data points. | 1. RMSSTD is easy to understand and calculate, requiring only basic mathematical operations.<br>2. It can be used for any type of clustering method.<br>3. It gives a single value that represents the overall quality of clustering.<br>4. It is sensitive to the size of the dataset, which makes it useful for comparing clustering results across datasets. | 1. RMSSTD can only be used for assessing internal validity, which means it cannot compare clustering results to an external ground truth.<br>2. It is not very robust to outliers and can be affected by noise in the data.<br>3. It does not consider the overall structure of the clustering result, only the spread of the points within each cluster. | Bioinformatics, Image Analysis, Natural Language Processing (NLP), Social Network Analysis, and Marketing. |
| 10 | R-Squared (RS), [88]. | Elbow | $O(n^2)$<br>Where: | 1. RS is based on a statistical measure of goodness-of-fit and | 1. RS requires a pre-specified number of clusters, which can be | Image Processing, Bioinformatics, |

| | | | | n refers to the data points. | is widely used as an internal validation index for clustering algorithms.<br>2. It is easy to calculate and interpret.<br>3. RS can measure the quality of clustering results and can be used to determine the optimal number of clusters.<br>4. RS is independent of cluster shapes and sizes, making it suitable for a wide range of clustering algorithms. | difficult to determine for large and complex datasets.<br>2. RS does not consider the relationship between the clusters and the underlying data distribution.<br>3. RS can produce different scores for different cluster partitions, which can make it difficult to select the best clustering result.<br>4. RS assumes that the data is normally distributed and may not perform well on non-normal datasets. | Customer Segmentation, and Pattern Recognition. |
|---|---|---|---|---|---|---|---|
| 11 | Standard Deviation Index (SD), [89]. | Min | O(n)<br>Where:<br>n refers to the data points. | 1. SD can reflect the distance between each cluster and its centroid.<br>2. SD can also compare the distances within the same cluster and between different clusters to measure the compactness and separation of the clusters. | 1. SD is sensitive to the size of the dataset. Therefore, it is not suitable for large datasets.<br>2. there may be some limitations in its accuracy and effectiveness. | | Pattern Recognition, Image Segmentation, Data Mining, Bioinformatics, Gene Expression, Protein Structure, and Medical Image Analysis. |
| 12 | Ray-Turi Index (RT), [90]. | Min | $O(n^2)$<br>Where:<br>n refers to the data points. | 1. RT does not require the true number of clusters as input<br>2. It is not sensitive to the shape and size of the cluster<br>3. It provides a simple and intuitive measure of clustering quality | 1. Computationally expensive for large datasets<br>2. It may fail to distinguish between similar clusters and noise<br>3. It assumes that the clusters are of equal size and shape, and have Gaussian distribution<br>4. It lacks theoretical justification compared to other clustering validity indices. | | Image Segmentation, Social Network Analysis, Pattern Recognition, Optimisation Problems, Feature Selection, and Gene Expression, Analysis. |
| 13 | S_Dbw Index, [77]. | Min | O(n)<br>Where:<br>n refers to the number of clusters. | 1. S_Dbw's ability to handle datasets with different sizes and shapes.<br>2. The ability to select the optimal number of clusters.<br>3. Less sensitive to | 1. S_Dbw has high computational complexity, sensitivity to outliers, and inability to handle datasets with overlapping clusters.<br>2. The suitability of S-Dbw as a clustering validity index heavily depends on the | | Image Segmentation, Sensor Networks, Social Network Analysis, Gene Expression, Analysis, |

| # | Index | Opt | Complexity | Advantages | Disadvantages | Applications |
|---|---|---|---|---|---|---|
| | | | | the initialization of the clustering algorithm compared to other indices. | characteristics of the dataset being analyzed, and therefore it may not always be the optimal choice for evaluating clustering algorithms. | Pattern Recognition, and Customer Segmentation. |
| 14 | I Index (I), [91]. | Max | $O(n^2)$ Where: n refers to the number of data points. | 1. Unlike other indices, the I-index is not sensitive to the number of clusters and is effective for datasets with varying numbers of clusters. 2. It provides a measure of both compactness and separation of clusters, making it a robust measure of cluster validity. 3. It is computationally efficient and can be easily implemented for large datasets with high dimensions. | 1. I-index assumes that the clusters are convex and isotropic, which may not always be true in real-world datasets. 2. It requires a large number of trials to find the optimal set of clustering parameters, which can be time-consuming. 3. It is not effective for datasets with skewed or imbalanced distributions, as it tends to produce similar scores for both well-clustered and poorly-clustered datasets. | Image Segmentation, Text Clustering, and Customer Segmentation in marketing analysis. |
| 15 | CS Index, [92]. | Max | $O(n^2)$ Where: n refers to the data points. | 1. The CS index is based on a simple and easily interpretable concept of cluster separability. 2. It can effectively evaluate clustering results and can handle different cluster shapes and sizes. 3. Its sensitivity to the clustering structure changes is low. | 1. CS-index has a high computational cost due to its $O(n^2)$ runtime complexity. 2. It may not be practical for very large datasets. 3. It is based on pairwise distances, which may not be ideal for high dimensional data. 4. Its effectiveness depends on the underlying dataset and may not be suitable for all types of data. | Pattern Recognition, Data Mining, Image Segmentation, Text Clustering, Bioinformatics, and Social Network Analysis. |
| 16 | PBM Index, [93]. | Max | $O(n^2)$ Where: n refers to the data points. | 1. The PBM Index is suitable for datasets with overlapping clusters or ambiguous cluster boundaries 2. It can handle datasets with noise 3. It does not require the number of clusters to be specified in advance | 1. The PBM Index is restricted to evaluating the accuracy of stochastic clustering algorithms -It relies on the availability of a true partition for comparison - It does not provide information about the appropriateness of the cluster solutions | Natural Language Processing (NLP), Image Processing Signal Processing, and Bioinformatics. |
| 17 | NIVA Index, [94]. | Min | $O(n^2)$ Where: n refers to the number of data points, | 1. It can handle both single and multiple cluster structures. 2. It can handle both numerical and nominal data. 3. It does not require any knowledge of the true class labels or optimal number of clusters. 4. It can be used to compare different clustering algorithms. 5. It is relatively easy to compute and implement. | 1. It is sensitive to the shape and size of the clusters which makes it less effective with complex and non-convex clusters. 2. It assumes that the clusters are spherical and evenly sized, which is not always true in real-world data. 3. It may not work well with highly imbalanced data. 4. It may not perform well with noisy or overlapping data. 5. It does not consider the interpretability of the clustering results. | Image Segmentation, Gene Expression Analysis, Natural Language Processing (NLP), and Customer Segmentation. |
| 18 | partitioning Davies-Bouldin index (PDBI), [95]. | 0 | The complexity of PDBI may depend on several factors, including the size of the dataset, the dimensionality of the data, and the specific implementation details. | It processes a relevant CVI even with complex data structures and noisy clusters. PDBI is deterministic, independent of clustering algorithms, and produces a normalized score between 0 and 1. | It has lacks to deal with very large/huge datasets and cope with high dimensional spaces. | Two dimensional benchmark data sets |

# Appendix B

Table B.1 depicts each internal clustering validity index that has been proposed after 2014 and compare them in terms of optimal value, runtime complexity, pros, cons and the applications that use those indices in their fields.

Table B.1: The summary of the recently proposed internal clustering validation methods

| No. | Index Name, Reference | Optimal Value | Runtime Complexity | Advantages | Disadvantages | Applications |
|---|---|---|---|---|---|---|
| 1 | Clustering Validation based on Nearest Neighbor (CVNN), [50]. | Min | $O(n^2)$ Where: n refers to the number of data points. | 1. The method is simple to implement. 2. It is useful for datasets with unknown underlying distributions. 3. It does not require labeled data. | 1. The method may produce inaccurate results if the nearest neighbors are not representative of the overall structure of the data. 2. The method may produce conflicting results for datasets with different densities. 3. The method is sensitive to parameter settings, and experimentally finding the optimal set of parameters is not trivial. | Image Segmentation, Bioinformatics, Social Network Analysis and Pattern Recognition. |
| 2 | WB Index, [96]. | Min | $O(n^2)$ Where: n refers to the number of data points. | 1. The method is easy to implement and computationally efficient. 2. It does not require any prior knowledge about the true clustering labels. 3. It works well for datasets that have different densities and different sizes of clusters. | 1. The method is sensitive to the initial configuration of the centroids and may result in suboptimal clustering. 2. It only considers the average distance between data points and the centroids and does not consider the shape and structure of the clusters. 3. It may not work well for datasets with high-dimensional features, such as text or image data, where the distance between data points becomes less meaningful. | Bioinformatics, Gene Expression Analysis and Protein Sequence Clustering. |
| 3 | KCE, [97]. | Max | $O(n)$ Where: n refers to the number of data points. | 1. KCE is easy to compute. 2. It is insensitive to the number of clusters. | 1. KCE is being affected by the initial configuration of clusters. 2. It is sensitive to the scale of the data. | Image Segmentation, Text Clustering and Customer Segmentation. |

| | | | | 3. It is capable of handling both balanced and unbalanced cluster sizes. | | |
|---|---|---|---|---|---|---|
| 4 | Representative Tree Index (RTI), [98]. | Max | O(n) Where: n refers to the number of data points. | 1. RTI is insensitive to the scale of the data. 2. It handles the problem of overlapping clusters. 3. It is able to generate a representative tree diagram of the data. | 1. RTI is sensitive to the presence of outliers and the initial configuration of the clusters. 2. It is complex and computationally intensive, and may not work well with large datasets. | Biology, Ecology and Genetics. |
| 5 | Compact-Separate Proportion Index (CSP), [99]. | Max | $O(n^2)$ Where: n refers to the number of data points. | 1. CSP takes into account both compactness and separation of clusters, making it more robust compared to other cluster validity indices. 2. It is able to handle datasets with different dimensions and non-isotropic shapes. | 1. CSP requires to set the number of clusters beforehand, which might not always be possible. 2. It does not adjust for the size of clusters, which could affect the results. 3. The computational cost of calculating CSP can be high for large datasets. | Image Segmentation, Remote Sensing and Pattern Recognition. |
| 6 | Density Estimation Index (DE), [100]. | Max | $O(n^2)$ Where: n refers to the number of data points. | 1. DE is robust to noise and can handle clusters of arbitrary shapes, including overlapping clusters. 2. It is less sensitive to cluster size and density variations compared to other clustering indices. | 1. DE may not be suitable for datasets with very sparse or dense clusters, as the density estimation can become unreliable in such cases. 2.The computational complexity of DE can also be a challenge for large datasets, as it requires the computation of pairwise distances between all data points. | Pattern Recognition, Data Mining and Image Processing. |
| 7 | Wemmert Gancarski Index (WG), [101]. | Max | $O(n^2)$ Where: n refers to the number of data points. | 1. WG is sensitive to density-based clustering methods, which can be useful for datasets with irregular shapes or varying densities. 2. It is that the index can be computed relatively quickly and | 1. WG may not be suitable for datasets with large disparities in cluster sizes, as it may place too much emphasis on larger clusters. 2. It assumes equal variances within clusters, which may not hold true for all | Pattern Recognition, Data Mining, Image Processing and Bioinformatics. |

| | | | | | | can handle large datasets. | datasets, leading to inaccurate results. | |
|---|---|---|---|---|---|---|---|---|
| 8 | Density-Core-based Clustering Validation Index (DCVI), [102]. | Min | $O(n)$ Where: n refers to the number of data points. | | | 1. DCVI provides a simple and efficient approach to evaluate the clustering quality by considering the density of the core members in a cluster. 2. It can handle both hierarchical and partitioning clustering techniques. 3. It does not require any prior knowledge about the data distribution or number of clusters. | 1. DCVI may fail to accurately measure the clustering quality in datasets that contain different density regions or irregular shaped clusters. 2. It is sensitive to the choice of the radius parameter, which defines the vicinity area for selecting the core members. 3. It may not be effective in large-scale datasets due to its high computational complexity. | Image Processing, Machine Learning, Data Mining and Clustering Analysis. |
| 9 | Clustering Validity Index based on Density-Involved Distance (CVDD), [103]. | Max | $O(n^2)$ Where: n refers to the number of data points. | | | 1. CVDD is computationally efficient compared to other validity indices, which makes it suitable for large datasets. 2. It can handle clusters of arbitrary shapes, sizes, and densities, which makes it more flexible than some other indices that assume spherical clusters. 3. It is more robust to noise and outliers than some other indices, as it considers the density of points in addition to their distances. | 1. CVDD requires specifying the kernel parameter, which can be challenging for some datasets. Choosing an inappropriate kernel parameter can affect the validity of the metric. 2. It may not perform well when the densities of the clusters are significantly different, as it assumes that all clusters have similar densities. 3. It may not be suitable for datasets with a high level of overlap between clusters, as it considers all points in the dataset, regardless of their true cluster assignment. | Pattern Recognition, Image Segmentation, Data Mining and Clustering Analysis. |
| 10 | Local Cores-based Cluster Validity Index (LCCV), [104]. | Max | $O(n^2)$ Where: n refers to the number of data points. | | | 1. Handling noise points effectively: LCCV can handle datasets with a | 1. Sensitivity to core density: LCCV is sensitive to the selection of core density, which can | Bioinformatics, Image Segmentation, Text Mining and |

| | | | | significant amount of noise points in them.
2. Identifying clusters with varying densities: LCCV can identify clusters with different densities, as it determines the optimal number of clusters by detecting points that form dense regions.
3. Computationally efficient: LCCV is computationally efficient because it uses a distance-based approach and does not require a pairwise distance matrix. | affect the performance of the algorithm.
2. Unable to handle clusters with irregular shapes or elongated clusters: LCCV may not be suitable for datasets with clusters that have complex shapes or elongated clusters.
3. Not suitable for determining the optimal number of clusters: LCCV is used to evaluate the quality of clustering results but it does not provide a definitive estimate of the optimal number of clusters in a dataset. | Social Network Analysis. |
|---|---|---|---|---|---|---|
| 11 | Distance-based Separability Measure (DSI), [105]. | Max | $O(n^2)$ Where: n refers to the number of data points. | 1. DSI is simple to calculate and does not require any assumptions about the underlying distribution of data.
2. It can handle both compact and non-compact clusters, making it useful for various applications.
3. It considers the distances between data points, which can be informative about how well-separated clusters are from each other. | 1. DSI is sensitive to the number of clusters and the size of the dataset, which can lead to overfitting or underfitting the data.
2. It is only applicable to numerical data, which limits its usefulness when working with categorical variables.
3. It does not provide a threshold for evaluating the quality of clusters, making it challenging to interpret the results in a meaningful way. | Bioinformatics, Image Segmentation, Text Mining and Social Network Analysis. |
| 12 | Synthetical Clustering Validity (SCV), [106]. | Max | $O(n^2)$ Where: n refers to the number of data points. | 1. SCV has the ability to evaluate clustering methods without relying on external criteria and the ability to compare the performance of different clustering algorithms on the same dataset. | 1. The assumption that the synthetic datasets accurately represent real-world datasets and the fact that it may not be suitable for all types of clustering algorithms.
2. It also requires significant computational resources to generate | Image Segmentation, Text Clustering, Gene Expression Analysis, Clustering of Webpages and Traffic Sign Recognition. |

| | | | | | | |
|---|---|---|---|---|---|---|
| | | | | 2. It also allows for the detection of overfitting and can provide insights into the optimal number of clusters for a dataset. | and evaluate synthetic datasets. | |
| 13 | Generalized Synthetical Clustering Validity (GSCV), [107]. | Max | $O(n^3)$ Where: n refers to the number of data points. | 1. GSCV has the ability to evaluate the quality of clustering results on more realistic datasets and the ability to identify more subtle differences between clustering algorithms. 2. It can also provide insights into the optimal number of clusters and the robustness of clustering algorithms. | 1. The increased computational resources required to generate and evaluate more complex datasets, which can be time-consuming. 2. The use of more complex datasets may lead to increased sensitivity to parameter settings, which can affect the validity of the results. 3. GSCV assumes that the synthetic datasets accurately represent real-world datasets, which may not always be the case. | Image Segmentation and Gene Expression Analysis. |
| 14 | Adjustment of within-cluster Distance (AWCD), [108]. | Max | $O(n^2)$ Where: n refers to the number of data points. | 1. It can capture the effects of the AWCD adjustment, which can lead to more accurate evaluation of clustering solutions compared to traditional validity indices. 2. It accounts for the impact of outliers or extreme values in the clustering results, which can improve the overall quality of cluster assignments. 3. It can be used with any type of data or distance metric and can provide meaningful insights into the structure and relationships within the data. | 1. AWCD requires knowledge of the optimal values for the adjustment parameters, which may not always be available or easy to estimate. 2. It may not always result in optimal clustering solutions, as the AWCD adjustment may not always reflect the underlying structure or relationships within the data. 3. It may be computationally intensive, especially for large datasets, which can increase the time and resources required for evaluating clustering solutions. | Image Segmentation, Document Clustering, and Customer Segmentation. |
| 15 | Contiguous Density Region | Max | $O(n)$ Where: | 1. CDR is a simple measure that is easy | 1. CDR is sensitive to the shape of the clusters, and might not | Bioinformatics, Image Segmentation, |
| | (CDR) Index, [109][110]. | | n refers to the number of data points. | to understand and calculate.<br>2. It can measure how compact and separate clusters are effectively.<br>3. It is less sensitive to noise compared to other validity indices.<br>4. It can be used to determine the optimum number of clusters for a given dataset.<br>5. It is less influenced by the number of clusters compared to other internal validity measures. | be useful when irregular clusters are present.<br>2. The CDR measure does not take into consideration the size of the clusters<br>3. Setting a threshold value for what constitutes a cluster can be subjective and based on the experience of the user<br>4. It is affected by the initialization of clusters and the choice of algorithm used. | and Pattern Recognition. |
| 16 | Validity Index for Arbitrary-Shaped Clusters based on the Kernel Density Estimation (VIASCKDE), [111]. | Max | $O(n^3)$<br>Where:<br>n refers to the number of data points. | 1. VIASCKDE can handle data with arbitrary shapes and densities, unlike traditional clustering methods that assume data is normally distributed.<br>2. It does not require any prior knowledge or parameters to be specified, which makes it easy to use and less prone to bias.<br>3. It is capable of detecting clusters of varying sizes, including dense and sparse regions. | 1. The method can be computationally expensive for large datasets, which can lead to slower processing times.<br>2. VIASCKDE is sensitive to the choice of kernel function and bandwidth, which can impact the accuracy of the clustering results.<br>3. The interpretation of the validity index score may be subjective, as it depends on the decision of what is considered a "good" or "bad" clustering result. | Anomaly Detection, Disease Clustering Analysis, Image Segmentation, Pattern Recognition, Bioinformatics and Spatial Analysis. |
| 17 | Augmented Non-shared Nearest Neighbors (ANCV), [52]. | Max | $O(n^2)$<br>Where:<br>n refers to the number of data points. | 1. It is a flexible and efficient method that can handle clusters of different sizes and shapes.<br>2. It handles noise and outliers well, by assigning them to a separate cluster.<br>3. It can be easily parallelized, which improves its scalability. | 1. It requires a large amount of memory since it computes an augmented covariance matrix for each cluster.<br>2. It may suffer from the curse of dimensionality, in high-dimensional spaces, where distances between points become less informative. | Medical Diagnosis, Image Segmentation, Customer Segmentation Data Analysis. |

| | | | | | 3. It does not perform well with highly overlapping clusters, as it can mistakenly assign some points to the wrong cluster. | |

## Appendix C

In this appendix, Table C.1 is represented for each external clustering validity index and compares them in terms of optimal value, runtime complexity, pros, cons, and the applications that use those indices in their fields.

Table C.1: The summary of the external clustering validation methods

| No. | Index Name, Reference | Optimal Value | Runtime Complexity | Advantages | Disadvantages | Applications |
|---|---|---|---|---|---|---|
| 1 | Purity (P), [112]. | Max | $O(n)$ Where: n refers to the number of data points. | 1. Purity is simple to understand and compute, making it easy to use. 2. It is suitable for evaluating the performance of clustering algorithms in large datasets. 3. It can be used to compare different clustering algorithms and select the one that produces the most homogeneous clusters. | 1. Purity only measures the homogeneity of each cluster, without considering the similarity between members of the same class. Hence, it may not capture the quality of clustering results accurately. 2. It may not be appropriate for datasets with unbalanced class distributions, as it may overestimate the quality of clustering results in such cases. 3. It does not take into account the overlap between clusters, which may result in poor quality clustering results. | Text Clustering, Image Segmentation and Customer Segmentation in Marketing. |
| 2 | Entropy (H), [113]. | Min | $O(n)$ Where: n refers to the number of data points. | 1. Entropy measures the degree of ambiguity between the true class labels and the assigned clusters in a clustering algorithm. 2. It is more robust to cluster overlapping and noise, as it considers all misclassifications instead of only non-overlaps. 3. It is easy to interpret, with higher entropy values indicating poor matches between clusters and classes. | 1. Entropy assumes that each true class should represent a separate cluster, which may not always be the case. 2. It does not consider the distribution of data points within each cluster, which may affect the visual quality of the clusters. 3. It may not be suitable for datasets with an unbalanced distribution of class sizes, as entropy values may be skewed towards the larger class. | Image Compression, Document Clustering, Machine Learning and Data Mining. |
| 3 | F-Measure (F), [114]. | Max | $O(n^2)$ Where: | 1. F-measure provides a single score that evaluates | 1. F-measure requires ground truth labels to | Image Segmentation, |

| | | | | n refers to the number of data points. | both precision and recall of a clustering algorithm.<br>2. It allows for comparison of different clustering algorithms based on a single performance metric.<br>3. It is suitable for datasets with imbalanced cluster sizes. | evaluate the clusters, which may not be available in unsupervised learning tasks.<br>2. It does not account for the overlap of clusters, which may be important in some clustering applications.<br>- Its performance may be affected by the choice of evaluation threshold. | Data Mining and Document Clustering. |
|---|---|---|---|---|---|---|---|
| 4 | Rand Index (R), [114]. | Max | $O(n^2)$<br>Where:<br>n refers to the number of data points. | 1. Rand Index is a simple and easy-to-understand metric that compares the similarity between two sets of clustering results.<br>2. It is suitable for evaluating the performance of clustering algorithms with any type of data (continuous, categorical, etc.).<br>3. It is not affected by the number of clusters or their sizes. | 1. Like F-measure, Rand Index requires ground truth labels for evaluation, which may not be available in various unsupervised clustering problems.<br>2. It does not account for chance agreement, which may be a problem when dealing with noisy data or when there is a high degree of similarity between different clusters.<br>3. It also suffers from a tendency to favor algorithms that produce clusters with similar sizes and number of components. | Bioinformatics, Image Segmentation, and Natural Language Processing (NLP). | |
| 5 | Jaccard Index (J), [115]. | Max | $O(n^2)$<br>Where:<br>n refers to the number of data points. | Jaccard Index is easily interpretable and applicable to all kinds of datasets. | 1. Jaccard Index is very sensitive to noise: small variations in the dataset can lead to significant changes in the resulting value.<br>2. It is biased towards clusters with similar sizes, meaning that clusters with low or high variability in size can be undervalued or overvalued, respectively.<br>3. It assumes that the number of clusters between the benchmark partition | Image Segmentation, Text Mining and Bioinformatics | |

| # | | | | | and the clustering algorithm is the same, which is often not the case in real-world applications. | |
|---|---|---|---|---|---|---|
| 6 | Normalized Mutual Information (NMI), [116]. | Max | O(n) Where: n refers to the number of data points. | 1. NMI ability to provide a normalized measure of the agreement between two clusters even if they have different numbers of clusters. 2. its independence from the particular clustering algorithm used. 3. It can handle categorical and numerical data. | 1. NMI is sensitive to noise and outliers, as well as its limited ability to detect local agreement within clusters. 2. It also requires a ground-truth clustering or a reference clustering to compare against, which is not always readily available in real-world applications. | Image Segmentation, Bioinformatics and Data Mining. |
| 7 | Min-Max Index (MMI), [117]. | Max | O(n) Where: n refers to the number of data points. | 1. The ability of MMI to handle overlapping clusters and its robustness to noise and outliers. 2. MMI does not require a ground-truth clustering or a reference clustering to compare against, making it more suitable for real-world applications. | 1. MMI is sensitive to the number of clusters and can penalize or favor clustering with larger or smaller number of clusters, respectively. 2. It is unable to distinguish between correct and wrong merges or splits of clusters. | Image Segmentation, Bioinformatics and Text Clustering |
| 8 | Fowlkes-Mallows Index (FM), [43,118]. | Max | $O(n^2)$ Where: n refers to the number of data points. | 1. FM is easy to understand and calculate. 2. It provides a single summary measure of clustering accuracy. 3. It is suited for datasets with known ground truth. | 1. FM does not take into account the cluster size or noise and assumes clusters to be of equal size and density. 2. It is sensitive to the number of clusters and can give misleading results if the number of clusters is not known beforehand. 3. It does not work well with overlapping or hierarchical clusters. | Bioinformatics, Social Network Analysis, Computer Vision and Natural Language Processing (NLP). |

## References

[1] B.A. Hassan, T.A. Rashid, H.K. Hamarashid, A novel cluster detection of COVID-19 patients and medical disease conditions using improved evolutionary clustering algorithm star, Comput Biol Med 138 (2021). https://doi.org/10.1016/j.compbiomed.2021.104866.

[2] B.A. Hassan, T.A. Rashid, A multidisciplinary ensemble algorithm for clustering heterogeneous datasets, Neural Comput Appl 33 (2021). https://doi.org/10.1007/s00521-020-05649-1.

[3] G. Gan, C. Ma, J. Wu, Data clustering: theory, algorithms, and applications, SIAM, 2020.

[4] M.R.M. Talabis, R. McPherson, I. Miyamoto, J.L. Martin, D. Kaye, Analytics Defined, in: Information Security Analytics, Elsevier, 2015: pp. 1–12. https://doi.org/10.1016/b978-0-12-800207-0.00001-0.

[5] K. Zakharov, Application of k-means clustering in psychological studies, Quant Method Psychol 12 (2016) 87–100. https://doi.org/10.20982/tqmp.12.2.p087.

[6] R. Röttger, Clustering of Biological Datasets in the Era of Big Data, J Integr Bioinform 13 (2016) 300. https://doi.org/10.2390/biecoll-jib-2016-300.

[7] N. M, A Comprehensive Overview of Clustering Algorithms in Pattern Recognition, IOSR J Comput Eng 4 (2012) 23–30. https://doi.org/10.9790/0661-0462330.

[8] P. Braun, A. Cuzzocrea, T.D. Keding, C.K. Leung, A.G.M. Padzor, D. Sayson, Game Data Mining: Clustering and Visualization of Online Game Data in Cyber-Physical Worlds, in: Procedia Comput Sci, Elsevier B.V., 2017: pp. 2259–2268. https://doi.org/10.1016/j.procs.2017.08.141.

[9] I. Zheliznyak, Z. Rybchak, I. Zavuschak, Analysis of clustering algorithms, in: Advances in Intelligent Systems and Computing, Springer Verlag, 2017: pp. 305–314. https://doi.org/10.1007/978-3-319-45991-2_21.

[10] B. Bohara, J. Bhuyan, F. Wu, J. Ding, A Survey on the Use of Data Clustering for Intrusion Detection System in Cybersecurity, International Journal of Network Security & Its Applications 12 (2020) 1–18. https://doi.org/10.5121/ijnsa.2020.12101.

[11] E. Zhu, R. Ma, An effective partitional clustering algorithm based on new clustering validity index, Appl Soft Comput 71 (2018) 608–621. https://doi.org/10.1016/J.ASOC.2018.07.026.

[12] F. Nielsen, Hierarchical Clustering, in: 2016: pp. 195–211. https://doi.org/10.1007/978-3-319-21903-5_8.

[13] R.J.G.B. Campello, P. Kröger, J. Sander, A. Zimek, Density-based clustering, Wiley Interdiscip Rev Data Min Knowl Discov 10 (2020). https://doi.org/10.1002/widm.1343.

[14] M. Tareq, E.A. Sundararajan, A. Harwood, A.A. Bakar, A Systematic Review of Density Grid-Based Clustering for Data Streams, IEEE Access 10 (2022) 579–596. https://doi.org/10.1109/ACCESS.2021.3134704.

[15] S.K. Choy, S.Y. Lam, K.W. Yu, W.Y. Lee, K.T. Leung, Fuzzy model-based clustering and its application in image segmentation, Pattern Recognit 68 (2017) 141–157. https://doi.org/10.1016/J.PATCOG.2017.03.009.

[16] S. Saitta, B. Raphael, I.F.C. Smith, A Bounded Index for Cluster Validity, Verlag Berlin Heidelberg, 2007.

[17] E. Horne, H. Tibble, A. Sheikh, A. Tsanas, Challenges of clustering multimodal clinical data: Review of applications in asthma subtyping, JMIR Med Inform 8 (2020). https://doi.org/10.2196/16452.

[18] A. Ghosal, A. Nandy, A.K. Das, S. Goswami, M. Panday, A Short Review on Different Clustering Techniques and Their Applications, in: Advances in Intelligent Systems and Computing, Springer Verlag, 2020: pp. 69–83. https://doi.org/10.1007/978-981-13-7403-6_9.

[19] X. Huang, L. Zhang, B. Wang, F. Li, Z. Zhang, Feature clustering based support vector machine recursive feature elimination for gene selection, Applied Intelligence 48 (2018) 594–607. https://doi.org/10.1007/s10489-017-0992-2.

[20]  T. Kinnunen, I. Sidoroff, M. Tuononen, P. Fränti, Comparison of clustering methods: A case study of text-independent speaker modeling, Pattern Recognit Lett 32 (2011) 1604–1617. https://doi.org/10.1016/j.patrec.2011.06.023.

[21]  P. Fränti, S. Sieranoja, How much can k-means be improved by using better initialization and repeats?, Pattern Recognit 93 (2019) 95–112. https://doi.org/10.1016/j.patcog.2019.04.014.

[22]  K. Krishna, M.N. Murty, Genetic K-means algorithm, IEEE Transactions on Systems, Man, and Cybernetics, Part B: Cybernetics 29 (1999) 433–439. https://doi.org/10.1109/3477.764879.

[23]  P. Fränti, Genetic algorithm with deterministic crossover for vector quantization, Pattern Recognit Lett 21 (2000) 61–68. https://doi.org/10.1016/S0167-8655(99)00133-6.

[24]  P. Fränti, Efficiency of random swap clustering, J Big Data 5 (2018). https://doi.org/10.1186/s40537-018-0122-y.

[25]  P. Fränti, J. Kivijärvi, Randomised Local Search Algorithm for the Clustering Problem, 2000.

[26]  S. Kalyani, K.S. Swarup, Particle swarm optimization based K-means clustering approach for security assessment in power systems, Expert Syst Appl 38 (2011) 10839–10846. https://doi.org/10.1016/J.ESWA.2011.02.086.

[27]  L. Bai, X. Cheng, J. Liang, H. Shen, Y. Guo, Fast density clustering strategies based on the k-means algorithm, Pattern Recognit 71 (2017) 375–386. https://doi.org/10.1016/J.PATCOG.2017.06.023.

[28]  John. Elder, ACM Digital Library., Association for Computing Machinery. Special Interest Group on Knowledge Discovery & Data Mining., Association for Computing Machinery. Special Interest Group on Management of Data., Fast approximate spectral clustering, (2009) 1406.

[29]  L. Morissette, S. Chartier, The k-means clustering technique: General considerations and implementation in Mathematica, Tutor Quant Methods Psychol 9 (2013) 15–24. https://doi.org/10.20982/tqmp.09.1.p015.

[30]  P. Fränti, S. Sieranoja, K-means properties on six clustering benchmark datasets, Applied Intelligence 48 (2018) 4743–4759. https://doi.org/10.1007/s10489-018-1238-7.

[31]  A. ben Said, R. Hadjidj, S. Foufou, Cluster validity index based on Jeffrey divergence, (2018). https://doi.org/10.1007/s10044-015-0453-7.

[32]  M.M. Elmorshedy, R. Fathalla, Y. El-Sonbaty, Feature Transformation Framework for Enhancing Compactness and Separability of Data Points in Feature Space for Small Datasets, Applied Sciences (Switzerland) 12 (2022). https://doi.org/10.3390/app12031713.

[33]  L. Jegatha Deborah, R. Baskaran, A. Kannan, A Survey on Internal Validity Measure for Cluster Validation, International Journal of Computer Science & Engineering Survey 1 (2010) 85–102. https://doi.org/10.5121/ijcses.2010.1207.

[34]  M. HALKID, Y. BATISTAKIS, M. VAZIRGIANNIS, On Clustering Validation Techniques, 2001.

[35]  B.A. Hassan, T.A. Rashid, A multidisciplinary ensemble algorithm for clustering heterogeneous datasets, Neural Comput Appl (2021). https://doi.org/10.1007/s00521-020-05649-1.

[36]  C. Chou, M. Su, E.L.-2nd W.Int.Conf. on Scientific, undefined 2002, Symmetry as a new measure for cluster validity, Researchgate.NetCH Chou, MC Su, E Lai2nd WSEAS Int. Conf. on Scientific Computation and Soft Computing, 2002•researchgate.Net (n.d.). https://www.researchgate.net/profile/Mu-Chun-Su/publication/255599197_Symmetry_as_A_new_Measure_for_Cluster_Validity/links/02e7e53a4289a0e26b000000/Symmetry-as-A-new-Measure-for-Cluster-Validity.pdf (accessed January 26, 2024).

[37]  Chow C.H, Su M.C and Lai Eugene. A new Validity Measure… - Google Scholar, (n.d.). https://scholar.google.co.uk/scholar?hl=en&as_sdt=0%2C5&q=Chow+C.H%2C+Su+M.C+and+Lai+Eugene.+A+new+Validity+Measure+for+Clusters+with+Different+Densities.+Pattern+Anal.+Applications%2C+7%2C+2004%2C+pp.2005-2020.&btnG= (accessed January 26, 2024).

[38] M. Halkidi, M. Vazirgiannis, V. Balislakis, Quality scheme assessment in the clustering process, Lecture Notes in Computer Science (Including Subseries Lecture Notes in Artificial Intelligence and Lecture Notes in Bioinformatics) 1910 (2000) 265–276. https://doi.org/10.1007/3-540-45372-5_26/COVER.

[39] S. Banerjee, A. Choudhary, S. Pal, Empirical Evaluation of K-Means, Bisecting K-Means, Fuzzy C-Means and Genetic K-Means Clustering Algorithms, n.d.

[40] T. Van Craenendonck, H. Blockeel, K.U. Leuven, Using Internal Validity Measures to Compare Clustering Algorithms, in: International Conference on Machine Learning, 2015: pp. 1–8.

[41] J. Hämäläinen, S. Jauhiainen, T. Kärkkäinen, Comparison of internal clustering validation indices for prototype-based clustering, Algorithms 10 (2017). https://doi.org/10.3390/a10030105.

[42] H. Meroufel, Comparative Study between Validity Indices to Obtain the Optimal Cluster, International Journal of Computer and Electrical Engineering 9 (2017) 343–350. https://doi.org/10.17706/IJCEE.2017.9.1.343-350.

[43] M.Z. Rodriguez, C.H. Comin, D. Casanova, O.M. Bruno, D.R. Amancio, L. da F. Costa, F.A. Rodrigues, Clustering algorithms: A comparative approach, PLoS One 14 (2019). https://doi.org/10.1371/journal.pone.0210236.

[44] M. Alhabo, L. Zhang, Multi-criteria handover using modified weighted TOPSIS methods for heterogeneous networks, IEEE Access 6 (2018) 40547–40558. https://doi.org/10.1109/ACCESS.2018.2846045.

[45] Q.M. Ashraf, M.H. Habaebi, M.R. Islam, TOPSIS-Based Service Arbitration for Autonomic Internet of Things, IEEE Access 4 (2016) 1313–1320. https://doi.org/10.1109/ACCESS.2016.2545741.

[46] D. Hooshyar, Y. Yang, M. Pedaste, Y.M. Huang, Clustering Algorithms in an Educational Context: An Automatic Comparative Approach, IEEE Access 8 (2020) 146994–147014. https://doi.org/10.1109/ACCESS.2020.3014948.

[47] D.B. Dias, R.C.B. Madeo, T. Rocha, H.H. Biscaro, S.M. Peres, Hand movement recognition for Brazilian Sign Language: A study using distance-based neural networks, in: Proceedings of International Joint Conference on Neural Networks, IEEE, Atlanta, GA, USA, 2009: pp. 697–704.

[48] S.I. Hassan, A. Samad, O. Ahmad, A. Alam, Partitioning and hierarchical based clustering: a comparative empirical assessment on internal and external indices, accuracy, and time, International Journal of Information Technology (Singapore) 12 (2020) 1377–1384. https://doi.org/10.1007/s41870-019-00406-7.

[49] A. Karanikola, C.M. Liapis, S. Kotsiantis, A comparative study of validity indices on estimating the optimal number of clusters, in: The 12th International Conference on Information, Intelligence, Systems and Applications, IEEE, Chania, Crete, Greece., 2021.

[50] Y. Liu, Z. Li, H. Xiong, X. Gao, J. Wu, S. Wu, Understanding and enhancement of internal clustering validation measures, IEEE Trans Cybern 43 (2013) 982–994. https://doi.org/10.1109/TSMCB.2012.2220543.

[51] O. Arbelaitz, I. Gurrutxaga, J. Muguerza, J.M. Pérez, I. Perona, An extensive comparative study of cluster validity indices, Pattern Recognit 46 (2013) 243–256. https://doi.org/10.1016/j.patcog.2012.07.021.

[52] X. Duan, Y. Ma, Y. Zhou, H. Huang, B. Wang, A novel cluster validity index based on augmented non-shared nearest neighbors, Expert Syst Appl 223 (2023). https://doi.org/10.1016/j.eswa.2023.119784.

[53] Marko Niemelä, Ä. Sami, K. Tommi, Comparison of Cluster Validation Indices with Missing Data, in: European Symposium on Artificial Neural Networks, Computational Intelligence and Machine Learning, 2018: pp. 461–466.

[54] Y. Lei, J.C. Bezdek, S. Romano, N.X. Vinh, J. Chan, J. Bailey, Ground truth bias in external cluster validity indices, Pattern Recognit 65 (2017) 58–70. https://doi.org/10.1016/j.patcog.2016.12.003.

[55] M. Rezaei, P. Franti, Can the Number of Clusters Be Determined by External Indices?, IEEE Access 8 (2020) 89239–89257. https://doi.org/10.1109/ACCESS.2020.2993295.

[56] M.C.P. De Souto, A.L. V Coelho, K. Faceli♯, T.C. Sakata♯, V. Bonadia♯, I.G. Costa, A comparison of external clustering evaluation indices in the context of imbalanced data sets, in: Brazilian Symposium on Neural Networks, IEEE, Curitiba, Brazil, 2012.

[57] B.A. Hassan, T.A. Rashid, Operational framework for recent advances in backtracking search optimisation algorithm: A systematic review and performance evaluation, Appl Math Comput (2019) 124919.

[58] B.A. Hassan, CSCF: a chaotic sine cosine firefly algorithm for practical application problems, Neural Comput Appl (2020) 1–20.

[59] B.A. Hassan, T.A. Rashid, Datasets on statistical analysis and performance evaluation of backtracking search optimisation algorithm compared with its counterpart algorithms, Data Brief 28 (2020) 105046.

[60] M.H.R. Saeed, B.A. Hassan, S.M. Qader, An Optimized Framework to Adopt Computer Laboratory Administrations for Operating System and Application Installations, Kurdistan Journal of Applied Research 2 (2017) 92–97.

[61] B.A. Hassan, A.M. Ahmed, S.A. Saeed, A.A. Saeed, Evaluating e-Government Services in Kurdistan Institution for Strategic Studies and Scientific Research Using the EGOVSAT Model, Kurdistan Journal of Applied Research 1 (2016) 1–7.

[62] B.A. Hassan, T.A. Rashid, S. Mirjalili, Formal context reduction in deriving concept hierarchies from corpora using adaptive evolutionary clustering algorithm star, Complex & Intelligent Systems (2021) 1–16.

[63] B.A. Hassan, T.A. Rashid, S. Mirjalili, Performance evaluation results of evolutionary clustering algorithm star for clustering heterogeneous datasets, Data Brief (2021) 107044.

[64] B.A. Hassan, S.M. Qader, A New Framework to Adopt Multidimensional Databases for Organizational Information System Strategies, ArXiv Preprint ArXiv:2105.08131 (2021).

[65] B.A. Hassan, Analysis for the overwhelming success of the web compared to microcosm and hyper-G systems, ArXiv Preprint ArXiv:2105.08057 (2021).

[66] B. Hassan, S. Dasmahapatra, Towards Semantic Web: Challenges and Needs, (n.d.).

[67] H. Hamarashid, S. Saeed, Usability Testing on Sulaimani Polytechnic University Website, Int. J. of Multidisciplinary and Current Research 5 (2017).

[68] S. Saeed, S. Nawroly, H.H. Rashid, N. Ali, Evaluating e-Court Services Using the Usability Tests Model Case Study: Civil Court Case Management, Kurdistan Journal of Applied Research 1 (2016) 76–83.

[69] H.K. Hamarashid, M.H.R. Saeed, S. Saeed, Designing a Smart Traffic Light Algorithm (HMS) Based on Modified Round Robin Algorithm, Kurdistan Journal of Applied Research 2 (2017) 27–30.

[70] H.K. Hamarashid, S.A. Saeed, T.A. Rashid, Next word prediction based on the N-gram model for Kurdish Sorani and Kurmanji, Neural Comput Appl 33 (2021) 4547–4566.

[71] H.K. Hamarashid, Utilizing Statistical Tests for Comparing Machine Learning Algorithms, (2021).

[72] B.A. Hassan, T.A. Rashid, H.K. Hamarashid, A Novel Cluster Detection of COVID-19 Patients and Medical Disease Conditions Using Improved Evolutionary Clustering Algorithm Star, Comput Biol Med (2021) 104866.

[73] B.B. Maaroof, T.A. Rashid, J.M. Abdulla, B.A. Hassan, A. Alsadoon, M. Mohamadi, M. Khishe, S. Mirjalili, Current Studies and Applications of Shuffled Frog Leaping Algorithm: A Review, Archives of Computational Methods in Engineering (2022) 1–16.

[74] E.C. Dalrymple-Alford, MEASUREMENT OF CLUSTERING IN FREE RECALL, 1970.

[75] J.C. Bezdek, M. Moshtaghi, T. Runkler, C. Leckie, The generalized c index for internal fuzzy cluster validity, IEEE Transactions on Fuzzy Systems 24 (2016) 1500–1512. https://doi.org/10.1109/TFUZZ.2016.2540063.

[76] F. Haouas, Z. Ben Dhiaf, A. Hammouda, B. Solaiman, A new efficient fuzzy cluster validity index: Application to images clustering, in: IEEE International Conference on Fuzzy Systems (FUZZ-IEEE), IEEE, Naples, 2017.

[77] M. Halkidi, M. Vazirgiannis, Clustering validity assessment: Finding the optimal partitioning of a data set, in: Proceedings - IEEE International Conference on Data Mining, ICDM, 2001: pp. 187–194. https://doi.org/10.1109/icdm.2001.989517.

[78]     J.C. Dunn, A fuzzy relative of the ISODATA process and its use in detecting compact well-separated clusters, Journal of Cybernetics 3 (1973) 32–57. https://doi.org/10.1080/01969727308546046.

[79]     T.C. Havens, J.C. Bezdek, J.M. Keller, M. Popescu, Dunn's Cluster Validity Index as a Contrast Measure of VAT Image, n.d.

[80]     T. Caliñski, J. Harabasz, A Dendrite Method Foe Cluster Analysis, Communications in Statistics 3 (1974) 1–27. https://doi.org/10.1080/03610927408827101.

[81]     D.L. Davies, D.W. Bouldin, A Cluster Separation Measure, IEEE Trans Pattern Anal Mach Intell PAMI-1 (1979) 224–227. https://doi.org/10.1109/TPAMI.1979.4766909.

[82]     J.C.R. Thomas, M.S. Peñas, M. Mora, New Version of Davies-Bouldin Index for Clustering Validation Based on Cylindrical Distance, in: Proceedings - International Conference of the Chilean Computer Science Society, SCCC, IEEE Computer Society, 2013: pp. 49–53. https://doi.org/10.1109/SCCC.2013.29.

[83]     L. Hubert, P. Arabie, Comparing Partitions, J Classif 2 (1985) 193–218.

[84]     A.E. Rafteryt, A Note on Bayes Factors for Log-linear Contingency Table Models with Vague Prior Information, 1986.

[85]     P.J. Rousseeuw, Silhouettes: a graphical aid to the interpretation and validation of cluster analysis, 1987.

[86]     X.L. Xie, G. Beni, A Validity Measure for Fuzzy Clustering, IEEE Transactions on Pattern Analysis and Machine Learning 13 (1991) 841–847.

[87]     L. Wilkinson, L. Engelman, J. Corter, M. Coward, Cluster Analysis, n.d.

[88]     M. Halkidi, Y. Batistakis, M. Vazirgiannis, Clustering Validity Checking Methods: Part II, n.d.

[89]     M. Halkidi, M. Vazirgiannis, Y. Batistakis, Quality Scheme Assessment in the Clustering Process, 2000.

[90]     S. Ray, R.H. Turi, Determination of Number of Clusters in K-Means Clustering and Application in Colour Image Segmentation, (2000).

[91]     U. Maulik, S. Bandyopadhyay, Performance Evaluation of Some Clustering Algorithms and Validity Indices, 2002. http://www.ics.uci.edu/.

[92]     C.H. Chou, M.C. Su, E. Lai, A new cluster validity measure and its application to image compression, Pattern Analysis and Applications 7 (2004) 205–220. https://doi.org/10.1007/s10044-004-0218-1.

[93]     M.K. Pakhira, S. Bandyopadhyay, U. Maulik, Validity index for crisp and fuzzy clusters, Pattern Recognit 37 (2004) 487–501. https://doi.org/10.1016/j.patcog.2003.06.005.

[94]     E. RENDÓN, R. GARCIA, I. ABUNDEZ, C. GUTIERREZ, E. GASCA, F. DEL RAZO, A. GONZALEZ, NIVA: A Robust Cluster Validity, in: New Aspects of Communications: Proceedings of the 12th WSEAS International Conference on Communications, WSEAS, 2008: pp. 241–248.

[95]     F. Ros, R. Riad, S. Guillaume, PDBI: A partitioning Davies-Bouldin index for clustering evaluation, Neurocomputing 528 (2023) 178–199. https://doi.org/10.1016/J.NEUCOM.2023.01.043.

[96]     Q. Zhao, P. Fränti, WB-index: A sum-of-squares based index for cluster validity, Data Knowl Eng 92 (2014) 77–89. https://doi.org/10.1016/j.datak.2014.07.008.

[97]     S. Jauhiainen, T. Kärkkäinen, A Simple Cluster Validation Index with Maximal Coverage, in: ESANN 2017 Proceedings, European Symposium on Artificial Neural Networks, Belgium, 2017: pp. 293–298. http://www.i6doc.com/en/.

[98]     J.C. Rojas-Thomas, M. Santos, M. Mora, New internal index for clustering validation based on graphs, Expert Syst Appl 86 (2017) 334–349. https://doi.org/10.1016/j.eswa.2017.06.003.

[99]     S. Zhou, Z. Xu, F. Liu, Method for Determining the Optimal Number of Clusters Based on Agglomerative Hierarchical Clustering, IEEE Trans Neural Netw Learn Syst 28 (2017) 3007–3017. https://doi.org/10.1109/TNNLS.2016.2608001.

[100] B.W. Sliverman, Density estimation for statistics and data analysis, 2018.

[101] M.B. Desgraupes, Package "clusterCrit" Type Package Title Clustering Indices Version 1.2.8, 2018. www.r-project.org.

[102] J. Xie, Z.Y. Xiong, Q.Z. Dai, X.X. Wang, Y.F. Zhang, A new internal index based on density core for clustering validation, Inf Sci (N Y) 506 (2020) 346–365. https://doi.org/10.1016/j.ins.2019.08.029.

[103] L. Hu, C. Zhong, An internal validity index based on density-involved distance, IEEE Access 7 (2019) 40038–40051. https://doi.org/10.1109/ACCESS.2019.2906949.

[104] D. Cheng, Q. Zhu, J. Huang, Q. Wu, L. Yang, A Novel Cluster Validity Index Based on Local Cores, IEEE Trans Neural Netw Learn Syst 30 (2019) 985–999. https://doi.org/10.1109/TNNLS.2018.2853710.

[105] S. Guan, M. Loew, An Internal Cluster Validity Index Using a Distance-based Separability Measure, in: Proceedings - International Conference on Tools with Artificial Intelligence, ICTAI, IEEE Computer Society, 2020: pp. 827–834. https://doi.org/10.1109/ICTAI50040.2020.00131.

[106] Q. Xu, Q. Zhang, J. Liu, B. Luo, Efficient synthetical clustering validity indexes for hierarchical clustering, Expert Syst Appl 151 (2020). https://doi.org/10.1016/j.eswa.2020.113367.

[107] Q. Xu, Q. Zhang, J. Liu, B. Luo, Efficient synthetical clustering validity indexes for hierarchical clustering, Expert Syst Appl 151 (2020) 113367. https://doi.org/10.1016/J.ESWA.2020.113367.

[108] Q. Li, S. Yue, Y. Wang, M. Ding, J. Li, A new cluster validity index based on the adjustment of within-cluster distance, IEEE Access 8 (2020) 202872–202885. https://doi.org/10.1109/ACCESS.2020.3036074.

[109] J.C. Rojas-Thomas, M. Santos, New internal clustering validation measure for contiguous arbitrary-shape clusters, International Journal of Intelligent Systems 36 (2021) 5506–5529. https://doi.org/10.1002/int.22521.

[110] C.T. Zahn, Graph-Theoretical Methods for Detecting and Describing Gestalt Clusters, 1971.

[111] A. Şenol, VIASCKDE Index: A Novel Internal Cluster Validity Index for Arbitrary-Shaped Clusters Based on the Kernel Density Estimation, Comput Intell Neurosci 2022 (2022). https://doi.org/10.1155/2022/4059302.

[112] Clustering Validation, n.d.

[113] N. Novoselova, I. Tom, Entropy-based cluster validation and estimation of the number of clusters in gene expression data, J Bioinform Comput Biol 10 (2012). https://doi.org/10.1142/S0219720012500114.

[114] K. Draszawka, J.S. Szyma´nski, External Validation Measures for Nested Clustering of Text Documents, n.d.

[115] D. Tsarev, M. Petrovskiy, I. Mashechkin, Supervised and Unsupervised Text Classification via Generic Summarization, 2013. www.mirlabs.net/ijcisim/index.html.

[116] S.B. Dalirsefat, A. da Silva Meyer, S.Z. Mirhoseini, Comparison of similarity coefficients used for cluster analysis with amplified fragment length polymorphism markers in the silkworm, Bombyx mori, Journal of Insect Science 9 (2009). https://doi.org/10.1673/031.009.7101.

[117] A. kumar Alok, S. Saha, A. Ekbal, A min-max Distance Based External Cluster Validity Index: MMI, in: 12th International Conference on Hybrid Intelligent Systems (HIS), IEEE, Pune, India, 2012: pp. 354–359.

[118] S. Wagner, D. Wagner, Comparing Clusterings-An Overview *, 2007.